\documentclass[10pt,twocolumn,letterpaper]{article}

\usepackage{iccv}
\usepackage{times}
\usepackage{epsfig}
\usepackage{graphicx}
\usepackage{amsmath}
\usepackage{amssymb}
\usepackage{booktabs}
\usepackage{caption}
\usepackage{multirow}
\usepackage{bbding}
\usepackage{pifont}
\usepackage{makecell}
\usepackage{gensymb}
\usepackage{capt-of}
\usepackage{subcaption}
\newcommand{\xmark}{\ding{55}}

\usepackage[accsupp]{axessibility}  

\usepackage[breaklinks=true,bookmarks=false]{hyperref}

\iccvfinalcopy 

\def\iccvPaperID{6338} 

\ificcvfinal\fi

\begin{document}

\title{MatrixCity: A Large-scale City Dataset \\ for City-scale Neural Rendering and Beyond}

\author{Yixuan Li$^1$$^*$ , Lihan Jiang$^2$$^*$ , Linning Xu$^1$, Yuanbo Xiangli$^1$\\ Zhenzhi Wang$^1$, Dahua Lin$^{1,2}$, Bo Dai$^2$\Envelope\\
	$^1$ The Chinese University of Hong Kong \quad
	$^2$ Shanghai AI Laboratory \\
	\tt\small 
        \{ly122,xl020,xy019,wz122,dhlin\}@ie.cuhk.edu.hk \\
        \tt\small 
        \{jianglihan, daibo\}@pjlab.org.cn
}


\twocolumn[{
    \renewcommand\twocolumn[1][]{#1}
    \maketitle
    \begin{center}
	\vspace{-3em}
	\includegraphics[width=0.998\linewidth]{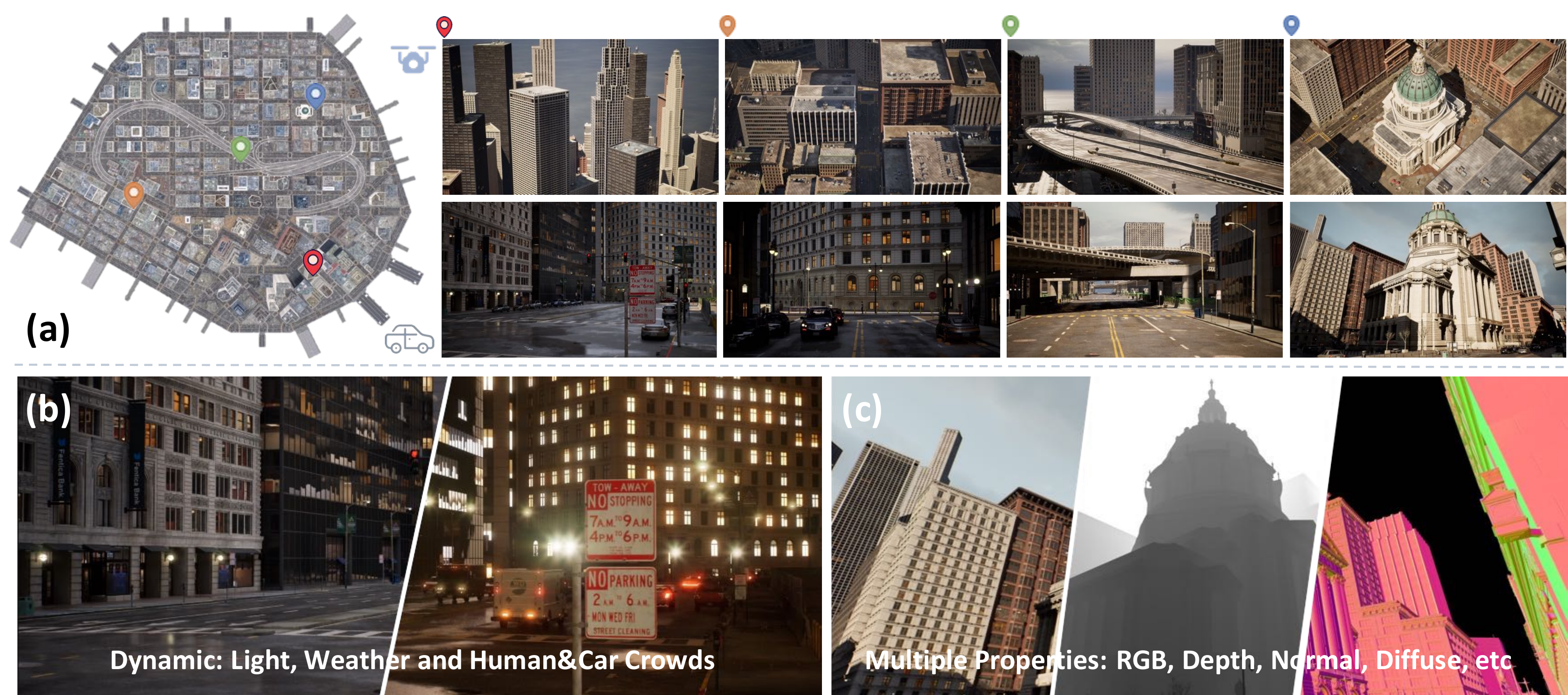}
	\captionof{figure}{Illustration of \emph{MatrixCity} dataset's \emph{Small City}. a) We collect high-quality large-scale city scene images for a city-scale neural rendering benchmark from Unreal Engine 5, capturing various city environments from multiple viewing angles. b) Our flexible environment control can collect data under dynamic environmental factors such as varying lighting and weather conditions. c) Our developed platform also facilitates the extraction of additional properties, including depth maps and normal maps. These features open a wide range of opportunities for future research in city-scale neural rendering.}\label{fig:overview}
    \end{center}
}]

\ificcvfinal\thispagestyle{empty}\fi

\begin{abstract}
Neural radiance fields (NeRF) and its subsequent variants have led to remarkable progress in neural rendering. While most of recent neural rendering works focus on objects and small-scale scenes, developing neural rendering methods for city-scale scenes is of great potential in many real-world applications. However, this line of research is impeded by the absence of a comprehensive and high-quality dataset, yet collecting such a dataset over real city-scale scenes is costly, sensitive, and technically difficult. To this end, we build a large-scale, comprehensive, and high-quality synthetic dataset for city-scale neural rendering researches. Leveraging the Unreal Engine 5 City Sample project, we develop a pipeline to easily collect aerial and street city views, accompanied by ground-truth camera poses and a range of additional data modalities. Flexible controls over environmental factors like light, weather, human and car crowd are also available in our pipeline, supporting the need of various tasks covering city-scale neural rendering and beyond. The resulting pilot dataset, MatrixCity, contains 67k aerial images and 452k street images from two city maps of total size $28km^2$. On top of MatrixCity, a thorough benchmark is also conducted, which not only reveals unique challenges of the task of city-scale neural rendering, but also highlights potential improvements for future works. The dataset and code will be publicly available at our project page:
\href{https://city-super.github.io/matrixcity/}{\textcolor{magenta}{https://city-super.github.io/matrixcity/}}.
\end{abstract}
\section{Introduction}
Realistic rendering of city-scale scenes is a crucial component of many real-world applications, including aerial surveying, virtual reality, film production, and gaming. 
While NeRF~\cite{DBLP:conf/eccv/MildenhallSTBRN20} has made notable advancements in rendering objects and small-scale scenes, only a few early attempts~\cite{DBLP:conf/cvpr/TancikCYPMSBK22,DBLP:conf/cvpr/TurkiRS22,DBLP:conf/eccv/XiangliXPZRTDL22} have sought to extend NeRF and its variants to larger city-scale scenes.
Due to the paucity of benchmark dataset, the complexity and challenges of city-scale neural rendering have not been thoroughly investigated.

Collecting a comprehensive and high-quality city-scale dataset in real-world is time consuming and resource intensive, and can be technically difficult. Moreover, it is also infeasible to control environmental factors, such as lighting conditions, weather patterns, and the presence of transient objects like pedestrians and vehicles. Thus, existing urban datasets~\cite{DBLP:conf/eccv/LinLHYXH22,DBLP:conf/cvpr/TurkiRS22,DBLP:journals/pami/CrandallOSH13} are limited to a few independent scenes rather than comprehensive city maps, failing to capture the diversity of urban environments. Furthermore, existing datasets often feature monotonous viewing angles, such as street-level~\cite{DBLP:conf/cvpr/TancikCYPMSBK22} or aerial imagery~\cite{DBLP:conf/eccv/LinLHYXH22,DBLP:conf/cvpr/TurkiRS22,DBLP:journals/pami/CrandallOSH13}, leading to partial city modeling with incomplete building geometries and ground-level details. Even if sufficient real-world city data is collected, legal or commercial issues can limit its accessibility, e.g., Block-NeRF dataset~\cite{DBLP:conf/cvpr/TancikCYPMSBK22} only provides access to 1km street data, and UrbanScene3D dataset~\cite{DBLP:conf/eccv/LinLHYXH22} offers only two real-world scenarios. Such restrictions significantly hinder the ability of researchers to advance the field of city-scale neural rendering.

This paper presents {\em MatrixCity}, a comprehensive and high-quality synthetic dataset to support the research of city-scale neural rendering as well as other extended tasks.
Specifically, {\em MatrixCity} has several distinguished features: 
1) \emph{High Quality.}  It is built in the City Sample project~\footnote{https://www.unrealengine.com/marketplace/product/city-sample} of Unreal Engine 5~\footnote{https://www.unrealengine.com/} with advanced graphic technologies which allows for the public release of rendered images\footnote{https://www.unrealengine.com/eula/unreal}. As shown in Figure~\ref{fig:overview}, this engine offers rich city details of fine-grained textures and geometries from its photo-realistic rendering quality with realistic lighting, shadow effects, and accurate ground-truth camera poses. 
2) \emph{Scale and Diversity.}
To create the MatrixCity dataset, we develop a plugin that can automatically capture data from the map of two cities provided by Unreal Engine 5, resulting in 172k and 347k images, respectively. These images cover areas equivalent to $2.7$km$^2$ and $25.3$km$^2$ in the real-world.
The captured regions showcase a broad spectrum of urban landscapes, mirroring the complexity and heterogeneity of genuine cities.
3) \emph{Controllable Environments.} 
Our developed plugin provides flexible control over a range of environmental factors that are uncontrollable in the real world, including lighting, weather, and human and car crowds. By decoupling these various factors, we are able to provide aligned data that can support in-depth research of city-scale neural rendering. 
4) \emph{Multiple Properties.} 
The plugin can also customize data collection trajectories, and extract multiple ground-truth components, including depth, normal and decomposed components of reflectance (e.g., diffuse, specular, metallic, etc.). Such advanced feature enables researchers to not only perform a range of city-scale neural rendering tasks under varying conditions but also supports other extended tasks, for example depth estimation and inverse rendering.

Our benchmark study demonstrates the value of MatrixCity in advancing city-scale neural rendering researches. We experiment with several state-of-the-art neural rendering methods to conduct empirical analyses first on aerial and street data respectively, then on the fused data from both modes. 
Preliminary results indicate that even with these advanced methods, city-scale neural rendering is still a far-reaching goal. Specifically, we identify several challenges: 1) In aerial data, learning high-rise city regions poses a greater challenge than low-rise/ground areas due to complex building structures and occlusion; 2) Street data contains significantly more details than aerial data, which raises challenges for model capacity. Although block-size aerial data modeling is feasible, modeling street data with the same size may be more difficult; 3) The view direction and level of details varied significantly between the two modes of data, making it difficult to train them together; 4) Current models generally perform poorly on smaller objects with more details and reflective buildings in urban scenes. These findings present significant opportunities to advance research in city-scale neural rendering.

In summary, our contributions are as follows:
\begin{itemize}
\item We construct a large-scale, high-quality dataset for city-scale neural rendering, named {\em MatrixCity}. 
This dataset emphasizes attributes pivotal to city-scale scenes, encompassing elements like dynamic interactions and lighting conditions.
MatrixCity contains both aerial and street-level images of complete city maps with extra depth, normal, and decomposed BRDF materials capable of supporting multiple tasks. 
\item We develop a plugin that leverages Unreal Engine 5 for automatic high-quality city data collection, allowing researchers to flexibly control lighting, weather, and transient objects. The plugin simplifies data collection for different task settings, making it a valuable tool for the community where users can build up advanced datasets as demanded. 
\item We conduct extensive studies on the MatrixCity dataset, which reveal some key challenges of city-scale neural rendering, and hopefully facilitate future research in this area

\end{itemize}

\section{Related work}
\subsection{3D Neural Representation at City Scale}
City-scale reconstruction has been studied for decades. Previous methods for representing geometry of a city mainly relied on raw point clouds acquired through either structure-from-motion ~\cite{DBLP:journals/cacm/AgarwalFSSCSS11} or Lidar sensors ~\cite{DBLP:journals/sensors/IlciT20}. Recently, with the emergence of Neural Radiance Fields (NeRF)~\cite{DBLP:conf/eccv/MildenhallSTBRN20}, novel view synthesis has become more efficient and effective. Numerous methods in this direction have further improved the speed~\cite{DBLP:conf/nips/LiuGLCT20,DBLP:conf/cvpr/0004SC22,DBLP:conf/cvpr/Fridovich-KeilY22,DBLP:journals/tog/MullerESK22,DBLP:conf/eccv/ChenXGYS22} and accuracy ~\cite{DBLP:journals/corr/abs-2010-07492,DBLP:conf/iccv/BarronMTHMS21,DBLP:conf/cvpr/BarronMVSH22} of reconstruction. NeRF is also used in a wide range of applications beyond novel view synthesis, such as inverse rendering~\cite{DBLP:conf/iccv/BossBJBLL21,DBLP:conf/cvpr/SrinivasanDZTMB21,DBLP:journals/tog/ZhangSDDFB21,DBLP:conf/eccv/RudnevESLGT22}, surface reconstruction~\cite{DBLP:conf/nips/WangLLTKW21,DBLP:journals/corr/abs-2208-12697,DBLP:conf/cvpr/AzinovicMGNT22,DBLP:conf/nips/YarivGKL21} or HDR synthesis~\cite{DBLP:conf/cvpr/HuangZFLWW22,DBLP:conf/eccv/Jun-SeongYYO22,DBLP:conf/cvpr/MildenhallHMSB22}. Although these methods demonstrate acceptable performance with small objects, grappling with urban scenes remains a significant challenge due to the limited representation capability of NeRF.

Based on these observations, recent methods are proposed for reconstructing radiance fields in urban-scale scenes. 
NeRF-W ~\cite{DBLP:conf/cvpr/Martin-BruallaR21} captured per-image appearance variations and separated the entire scene into static and transient components, enabling the modeling of unstructured collections of in-the-wild photographs. Block-NeRF ~\cite{DBLP:conf/cvpr/TancikCYPMSBK22} extends NeRF-W ~\cite{DBLP:conf/cvpr/Martin-BruallaR21} to model an neighborhood of San Francisco by dividing up urban environments into individually small Block-NeRFs. Mega-NeRF ~\cite{DBLP:conf/cvpr/TurkiRS22} also adopts the advantages of NeRF-W ~\cite{DBLP:conf/cvpr/Martin-BruallaR21} and Block-NeRF ~\cite{DBLP:conf/cvpr/TancikCYPMSBK22} by first decomposing large-scale fly-view scenes into small spatial cells and then training these cells in parallel. Urban Radiance Fields ~\cite{DBLP:conf/cvpr/RematasLSBTFF22} synthesizes novel RGB images and extracted 3D surfaces from a combination of panoramas and Lidar inputs in the urban environments. BungeeNeRF ~\cite{DBLP:conf/eccv/XiangliXPZRTDL22} introduces progressive modeling with multi-level supervision to handle city-scale data with varying levels of detail. Despite the progress made by the aforementioned methods~\cite{DBLP:conf/cvpr/Martin-BruallaR21,DBLP:conf/cvpr/TancikCYPMSBK22,DBLP:conf/cvpr/TurkiRS22,DBLP:conf/cvpr/RematasLSBTFF22,DBLP:conf/eccv/XiangliXPZRTDL22}, there is no unified dataset for evaluating these methods due to their varying settings. Significant challenge still remains in the city reconstruction problem, particularly in integrating aerial data and street-level data with varying levels of detail.

\subsection{NeRF-based Datasets and Benchmarks}
Several benchmarks based on NeRF are proposed in the recent two years, which focus on the effective and better reconstruction of single objects ~\cite{DBLP:conf/eccv/MildenhallSTBRN20, DBLP:conf/nips/LiuGLCT20, DBLP:conf/cvpr/JensenDVTA14}, indoor scenes ~\cite{DBLP:conf/cvpr/DaiCSHFN17}, or outdoor unbounded scenes~\cite{DBLP:conf/cvpr/BarronMVSH22, DBLP:journals/tog/KnapitschPZK17}. While there have been some good attempts to collect high-quality large-scale datasets using high-precision acquisition equipment \cite{DBLP:journals/pami/CrandallOSH13, DBLP:conf/eccv/LinLHYXH22, DBLP:conf/cvpr/TancikCYPMSBK22, DBLP:journals/corr/abs-2301-06782} as shown in Table~\ref{tab:dataset_comparsion}, the high acquisition costs limit their size and scale. Some datasets are limited to only a few independent scenes that are far from urban-scale or are not fully open-source due to privacy and commercial reasons. For instance, Mill 19 dataset~\cite{DBLP:conf/cvpr/TurkiRS22} only includes two suburban-like scenes, and the Quak 6D~\cite{DBLP:journals/pami/CrandallOSH13} and OMMO~\cite{DBLP:journals/corr/abs-2301-06782} datasets focus on a limited number of independent scenes that are not city-scale. Waymo Block-NeRF~\cite{DBLP:conf/cvpr/TancikCYPMSBK22} dataset only grants access to 100 seconds of driving data and Urban Scene3D dataset~\cite{DBLP:conf/eccv/LinLHYXH22} only releases two real-world scenarios.  
Additionally, existing real-world datasets commonly provide only one type of image data, such as street-level or aerial imagery, which makes modeling buildings incomplete \cite{DBLP:conf/eccv/LinLHYXH22, DBLP:journals/corr/abs-2301-06782}. Collecting real data in outdoor scenes poses significant challenges due to difficulties in controlling environmental factors such as pedestrian movement, weather, and lighting. As a result, a standard and comprehensive benchmark for city-scale neural rendering has not yet been established. Existing outdoor NeRF-based benchmarks like OMMO~\cite{DBLP:journals/corr/abs-2301-06782} are too trivial to explore and analyze the urban implicit scene representation. To address these issues, we develop a plugin in Unreal Engine 5 to easily collect aerial and street city data with ground-truth camera poses. We built a city-scale and multitasking dataset that includes both fly-view and street-view images and propose a new city-scale benchmark for neural rendering. We also provide a detailed analysis of the challenges and opportunities of NeRF in urban environments.

\section{MatrixCity Dataset}
\begin{figure*}[t]
\begin{center}
\includegraphics[width=1.0\linewidth]{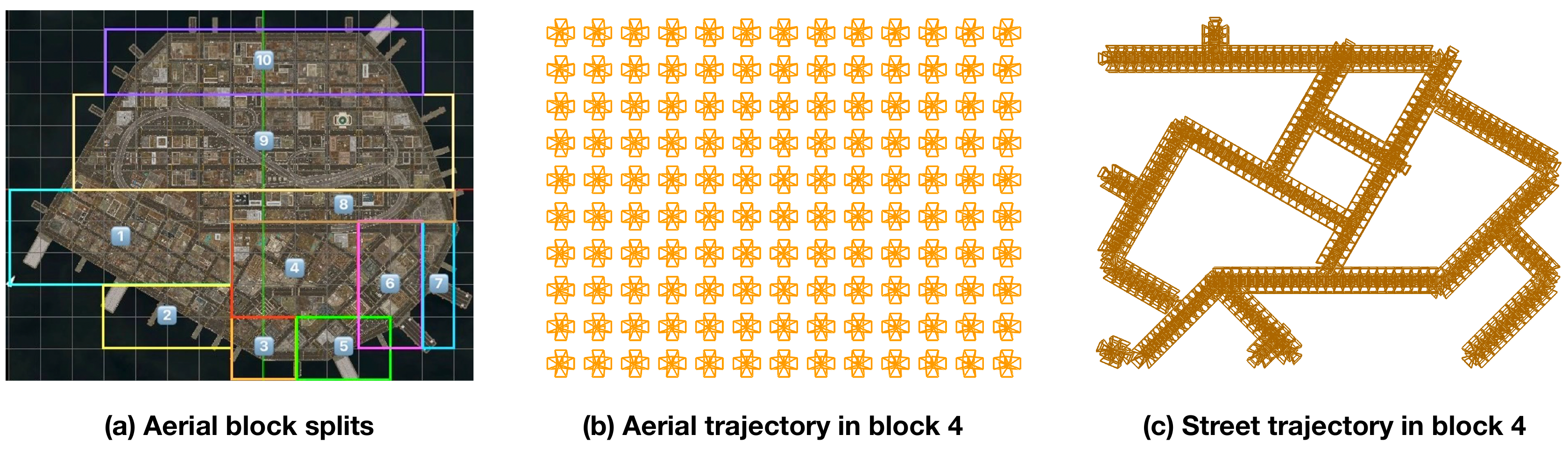}
\end{center}
\caption{Illustration of data collection in the \emph{small city} in Unreal Engine 5. (a) Aerial block split for the entire \emph{small city}; (b\&c) Camera aerial and street trajectory of block 4 (visualized in bird-eye views) used in our plugin for data collection.}
\label{fig:block}
\end{figure*}

The MatrixCity dataset aims to introduce a new challenging benchmark to the field of city-scale neural rendering by providing comprehensive city maps consisting of both aerial and street-level data. In addition to RGB images, we also offer normal, depth, and decomposed reflectance properties to support other tasks. Moreover, we can flexibly control environmental factors, including light direction and intensity, fog density, and human or vehicle crowding, to enable simulating real-world dynamic situations. Sec~\ref{sec:data_construct} describes our data construction procedure. Sec~\ref{sec:data_stat} and ~\ref{sec:data_character} provide detailed statistics and characteristics of this dataset.

\subsection{Dataset Construction}
\label{sec:data_construct}

\smallskip
\noindent\textbf{City Data Collection.} 
Densely captured 2D images with sufficient multi-view supervision are required to learn a faithful scene geometry, especially for large city scenes. Collecting a sufficient amount of data in Unreal Engine 5 for city scenes is a complex process that requires adjusting camera trajectories to capture specific viewpoints. Although the Unreal Engine 5 offers a movie render queue plugin for high-quality image rendering, it can be time-consuming and inflexible to manually set up the position, rotation, and frame number of key points. For urban settings, it is not practical to manually set camera trajectories in a city-scale environment. To address this, we develop a plugin that automatically generates camera trajectories, reducing the need for manual annotation and increasing the efficiency of data collection. The developed plugins can also be used in other Unreal Engine projects.

For \textbf{aerial-view} collection, we divide the city map into 10 blocks based on building heights (Figure~\ref{fig:block} (a)) to better capture the building details. We provide the height of every collected block in the Table~\ref{tab:high_block}. Note that current neural scene representations are generally suitable for bounded scenes, where scenes with large variation in height may cast great difficulty for accurate ray sampling. We then generate trajectories using the input of camera height and the four vertices' coordinates of the corresponding block (Figure~\ref{fig:block} (b))
Our plugin puts four cameras at each capturing location, with each camera rotating $90\degree$ apart from each other in the yaw direction and identical pitch values. The pitch value for the floor area is $-45\degree$, while it is $-60\degree$ for high-rise area as there are more occlusions at higher levels.

For \textbf{street-view} data collection, we manually annotate the start and end points of each road and use them as inputs to generate straight-line trajectories with our plugin. We position six perspective cameras at each capturing location to render a cube map, providing a comprehensive view of the surroundings. Note that the cube map can be naturally transformed into panorama images, which are suitable for capturing the street views as much as possible with limited camera positions. Figure~\ref{fig:block} (c) shows the resulting street-level trajectories for a specific block. Our plugin saves the generated camera trajectories as sequence assets of Unreal Engine, which can be easily reused to render images with different environmental settings. We will enhance our plugin to support more complex camera trajectories in the future, enabling us to generate even higher quality city-scale data. Note that we adapt auto-exposure to collect data. If we use the same fixed exposure for two types of data, the street views will be under-exposured while the aerial views will be over-exposured. HDR images will be included in the future.

\smallskip
\noindent\textbf{Quality Control.} To build a high-quality dataset for city-scale neural rendering, we utilize several mechanisms to ensure that the rendered images are of high quality and that the camera poses are accurate. Rather than using the more efficient real-time rendering pipeline, which often produces flickering images, we use the movie render queue plugin to render images with movie-level standards. Additionally, we set the Engine Scalability Settings to the best, turn off the motion blur and use anti-aliasing during the rendering process to achieve the highest possible image quality. We inspect the images thoroughly after rendering to remove any aerial views that look outside the map boundaries and ensure that there are no object clippings. Unreal Engine 5 provides ground-truth camera poses, which we have further verified through additional experiments to ensure their accuracy. Even with a small set of street data, training the MipNeRF-360~\cite{DBLP:conf/cvpr/BarronMVSH22} model yields almost perfect novel view synthesis results, as demonstrated in Figure~\ref{fig:visual_road}. This confirms the accurate annotation of our camera poses. Overall, by adopting these mechanisms, we ensure that the MatrixCity dataset provides high-quality images with precise camera poses, which is crucial for city-scale neural rendering research. Without considering noises like inaccurate pose and motion blur, we intend to gain more insights about the intrinsic challenges of city scenes since isolating these noises from real data is generally infeasible.

\begin{figure*}[t]
\begin{center}
\includegraphics[width=1.0\textwidth]{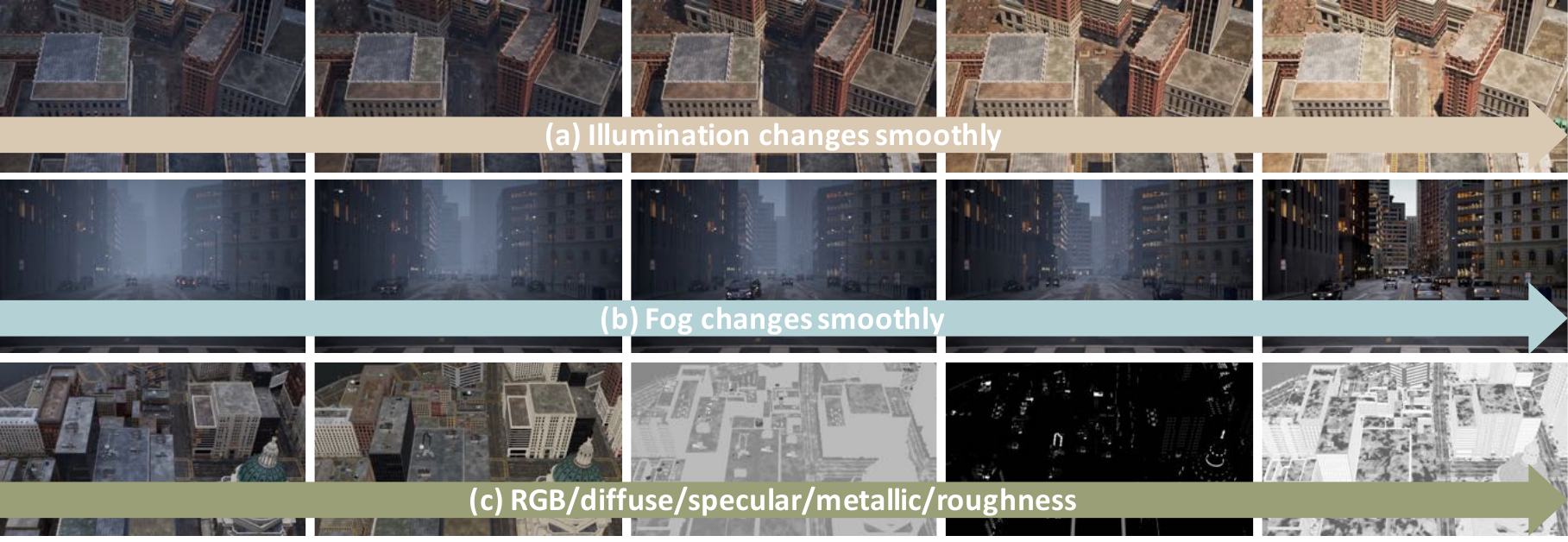}
\end{center}
   \caption{Illustration of controlling dynamic environment factors in Unreal Engine 5 such as illumination (a), fog density (b) and decomposed reflectance (c). }
\label{fig:illumination&fog}
\end{figure*}

\begin{figure}[h]
\begin{center}
\includegraphics[width=0.5\textwidth]{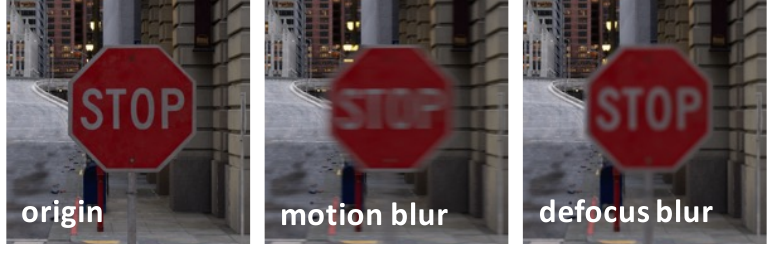}
\end{center}
\caption{Examples of camera motion blur and defocus blur.}
\label{fig:blur}
\end{figure}

\smallskip
\noindent\textbf{Dynamic Environments.} The City Sample project of Unreal Engine 5 provides a plethora of powerful functions that allow for the creation of dynamic city scenes. As shown in Figure~\ref{fig:overview}, we have the ability to control the presence of moving people and cars in the scene, adding to the realism of the environment. Additionally, we can quantitatively adjust the angle and intensity of the lighting to emulate the natural changes in light throughout a day, as demonstrated in Figure~\ref{fig:illumination&fog}(a). We can also control the amount of fog in a scene, as shown in Figure~\ref{fig:illumination&fog}(b), providing another quantitative tool for enhancing realism. Taken together, these functions allow for the simulation of almost all basic dynamic situations found in the real world. In addition, general camera noises like motion blur and defocus blur shown in Figure~\ref{fig:blur} can be simulated in Unreal Engine. Such varying lighting, weather conditions, moving objects and camera noises will lead to more realistic and accurate city-scale neural rendering. 

\smallskip
\noindent\textbf{Multiple Properties.} Figure~\ref{fig:overview} (c) and Figure~\ref{fig:illumination&fog} (c) illustrates the various intermediate products generated by Unreal Engine during the rendering process, including depth, normal, and decomposed components (diffuse, specular, metallic, and roughness). These attributes are especially important for studies on inverse rendering and semantic analysis, which are popular for city scene analysis. Our plugin offers the ability to extract these properties without incurring any additional costs, which can be prohibitively expensive to obtain in real-world scenarios.

\subsection{Dataset Statistics}
\label{sec:data_stat}
\begin{table*}[htbp]
\small
  \centering
  \resizebox{\linewidth}{!}{
    \begin{tabular}{ccccccccc}
    \toprule
    Dataset & \multicolumn{1}{l}{\#Images} & Level & Types & Source & Lighting & Human/Car & Weather &   D-Reflectance  \\ \midrule
    UrbanScene3D~\cite{DBLP:conf/eccv/LinLHYXH22} & \multicolumn{1}{l}{128K} & Scene & Aerial & Synthetic \& Real&   \xmark       &   \xmark    &   \xmark    & \xmark \\
    Quad 6K~\cite{DBLP:journals/pami/CrandallOSH13} & \multicolumn{1}{l}{5.1K} & Scene & Aerial & Real &      \xmark       &    \xmark   &   \xmark   &  \xmark\\
    Mill 19~\cite{DBLP:conf/cvpr/TurkiRS22} & \multicolumn{1}{l}{3.6K} & Scene & Aerial &   Real & \xmark   &      \xmark        &  \xmark     &  \xmark\\
    Waymo Block-NeRF~\cite{DBLP:conf/cvpr/TancikCYPMSBK22} & \multicolumn{1}{l}{12K} & City & Street &Real & $\checkmark$            &    \xmark   &    \xmark   & \xmark \\
    OMMO~\cite{DBLP:journals/corr/abs-2301-06782}  & \multicolumn{1}{l}{14.7K} & Scene & Aerial &  Real & $\checkmark$             &   \xmark    &    \xmark   &  \xmark\\ 
    KITTI-360~\cite{liao2022kitti} & \multicolumn{1}{l}{300K} & City & Street & Real &    \xmark   &      $\checkmark$        &  \xmark     &  \xmark\\
    NuScenes~\cite{caesar2020nuscenes} & \multicolumn{1}{l}{1.4M} & City & Street & Real &  \checkmark             &    $\checkmark$   &    \checkmark   & \xmark \\
    Waymo Open~\cite{sun2020scalability} & \multicolumn{1}{l}{1M} & City & Street & Real &  \checkmark           &   $\checkmark$    &    \checkmark   &  \xmark  \\ \bottomrule\midrule
    \textbf{Ours}  &   \multicolumn{1}{l}{519k}  & City  & Aerial+Street & Synthetic & $\checkmark$      & $\checkmark$   & $\checkmark$         & $\checkmark$  \\ \bottomrule
    \end{tabular}%
    }
    \caption{Comparison of statistics and properties between our {\em MatrixCity} dataset with previous datasets.}
    \vspace{5pt}
    \label{tab:dataset_comparsion}
\end{table*}%

The MatrixCity dataset comprises two scenes from the City Sample project: Small City covering an area of $2.7km^2$ and Big City spanning $25.3km^2$. In total, we collect 67k aerial images and 452k street-level images to ensure comprehensive coverage. As shown in Table~\ref{tab:dataset_comparsion}, many of current datasets~\cite{DBLP:conf/eccv/LinLHYXH22,DBLP:journals/pami/CrandallOSH13,DBLP:journals/corr/abs-2301-06782} do not offer the dense image captures of the whole city but small-size independent scenes. Although Waymo Block-NeRF~\cite{DBLP:conf/cvpr/TancikCYPMSBK22} dataset densely collects an area of approximately $960m \times 570m$, it only contains street data and results in the reconstructed buildings incomplete. All the existing datasets do not have quantitatively controllable environments including light, weather and human and car crowds, nor multiple properties like normal, depth, the decomposed reflectance components, etc, which restricts the in-depth study of city-scale neural rendering in dynamic scenes and other extension tasks. KITTI-360~\cite{liao2022kitti}, NuScenes~\cite{caesar2020nuscenes}, Waymo Open~\cite{sun2020scalability} are not designed for neural rendering purposes and only provide limited camera viewpoints. 

\subsection{Dataset Characteristics}
\label{sec:data_character}
\smallskip
\noindent\textbf{High Quality.} For constructing MatrixCity dataset, we use the City Sample project and the movie-level plugin named movie render queue of the Unreal Engine 5 which is demonstrated to reproduce \emph{The Matrix Awakens}. Unlike games, the rendering process is not real-time and costs huge computations with pre-defined camera poses. Such movie-level rendering quality enables the collection of realistic city-scale data similar to the real world with fully dynamic environment factors. Vigorous quality control is performed during its collection phase.

\begin{table*}[htbp]
\label{tab:high_benchmark}
\centering
\resizebox{\linewidth}{!}{
\begin{tabular}{c|ccc|ccc|ccc|ccc|ccc|c}
\toprule
                                    & \multicolumn{3}{c|}{ \textbf{NeRF} ~\cite{DBLP:conf/eccv/MildenhallSTBRN20}}                                            & \multicolumn{3}{c|}{\textbf{DVGO} ~\cite{DBLP:conf/cvpr/0004SC22}}                                            & \multicolumn{3}{c|}{\textbf{TensoRF} ~\cite{DBLP:conf/eccv/ChenXGYS22}}                                                                & \multicolumn{3}{c|}{\textbf{Instant-NGP} ~\cite{DBLP:journals/tog/MullerESK22}} & \multicolumn{3}{c|}{\textbf{MipNeRF-360} ~\cite{DBLP:conf/cvpr/BarronMVSH22}} \\ 
\multirow{-2}{*}{Block}           & PSNR $\uparrow$                        & SSIM  $\uparrow$                       & LPIPS  $\downarrow$                      & PSNR  $\uparrow$                       & SSIM $\uparrow$                        & LPIPS $\downarrow$                       & PSNR  $\uparrow$                       & SSIM  $\uparrow$                       & LPIPS $\downarrow$                       & PSNR  $\uparrow$                           & SSIM $\uparrow$    & LPIPS $\downarrow$   & PSNR $\uparrow$            & SSIM $\uparrow$            & LPIPS  $\downarrow$ & \multirow{-2}{*}{\makecell[c]{Average \\ Height}}         \\ \midrule
{Block\_A}   & 23.15 & 0.561 & 0.649 & 25.04 & 0.677 & 0.520 & {25.96} & {0.720} & {0.462} & \textbf{27.21}     & \textbf{0.793}    & \textbf{0.376}    & \underline{26.64}            & \underline{0.772}            & \underline{0.406}    & 150        \\
{Block\_B}  & 22.94                        & 0.613 & 0.485 & 22.72 & 0.649 & 0.463 & \underline{24.95} & \underline{0.776} & \underline{0.326} & \textbf{25.45}     & \textbf{0.826}    & \textbf{0.271}    & {24.80}            & {0.765}            & {0.352}    & 432        \\
{Block\_C} & 22.15                        & 0.590 & 0.527 & 21.39 & 0.649 & 0.475 & \underline{24.11} & {0.754} & {0.370} & 23.21     &   \textbf{0.788}      &    \textbf{0.311}      & \textbf{24.20}            & \underline{0.759}            & \underline{0.365}     & 419       \\
{Block\_D}     & 23.09                        & 0.570 & 0.548 & 24.14 & 0.656 & 0.486 & {24.99} & {0.712} & {0.416} & \underline{26.24}     & \underline{0.785}    & \underline{0.338}    & \textbf{26.45}            & \textbf{0.790}            & \textbf{0.338}      & 250      \\
{Block\_E}    & 23.53                        & 0.612 & 0.534 & 24.74 & 0.704 & 0.467 & {25.66} & {0.749} & {0.408} & \underline{26.36}     & \underline{0.807}    & \textbf{0.335}    & \textbf{26.54}            & \underline{0.811}            & \underline{0.338}     & 200       \\
{Overall}         & 22.97 & 0.589 & 0.548 & 23.61 & 0.667 & 0.482 & {25.13} & {0.762} & {0.396} & \underline{25.69}         &    \textbf{0.800}      &     \textbf{0.326}     & \textbf{25.73}            & \underline{0.779}            & \underline{0.360}   & 279\\
\bottomrule
\end{tabular}
}
\caption{Performance comparison of representative neural rendering methods on the aerial data of our {\em MatrixCity} benchmark.}
 \label{tab:high_results}%
\end{table*}
\smallskip
\noindent\textbf{Large-scale and Diversity.} The City Sample project of Unreal Engine 5 includes two cities with a large-scale coverage, which captures varying buildings, pedestrians, signs, vehicles, and lighting conditions, resulting in more diverse and realistic outdoor scenes that are representative of real-world cities. This ensures that researchers have access to a broad range of data to train their models on, leading to more accurate and effective city-scale neural rendering.

\smallskip
\noindent\textbf{Controllable Environments.} Unlike the real world data, we could control the lighting angle and intensity, the density and height of fog, and the density of flow of pedestrians and vehicles in a fine-grained manner. This flexibility enables us to generate dynamic scenarios of city scenes that would be difficult to capture in real-world data. This level of control over the environment allows for more detailed exploration of how different factors influence the training process of city-scale neural rendering.

\smallskip
\noindent\textbf{Multiple Properties.} Our developed plugin is able to extract additional information such as depth, normal and the decomposed reflectance components with minimum extra cost in Unreal Engine 5. This information supports additional tasks such as depth estimation, inverse rendering, which cannot be supported by real-world data without excessive labor.

\smallskip
\noindent\textbf{Applications.}
By exploring neural rendering models on MatrixCity, we can transfer the algorithms to real-world urban scenes.
This may facilitate the creation of scenes for applications ranging from video games and virtual reality to autonomous driving and virtual studios. Additionally, these rendered environments can enable seamless interactions with digital humans within the metaverse.

\begin{figure*}[htbp]
\begin{center}
\includegraphics[width=1.0\textwidth]{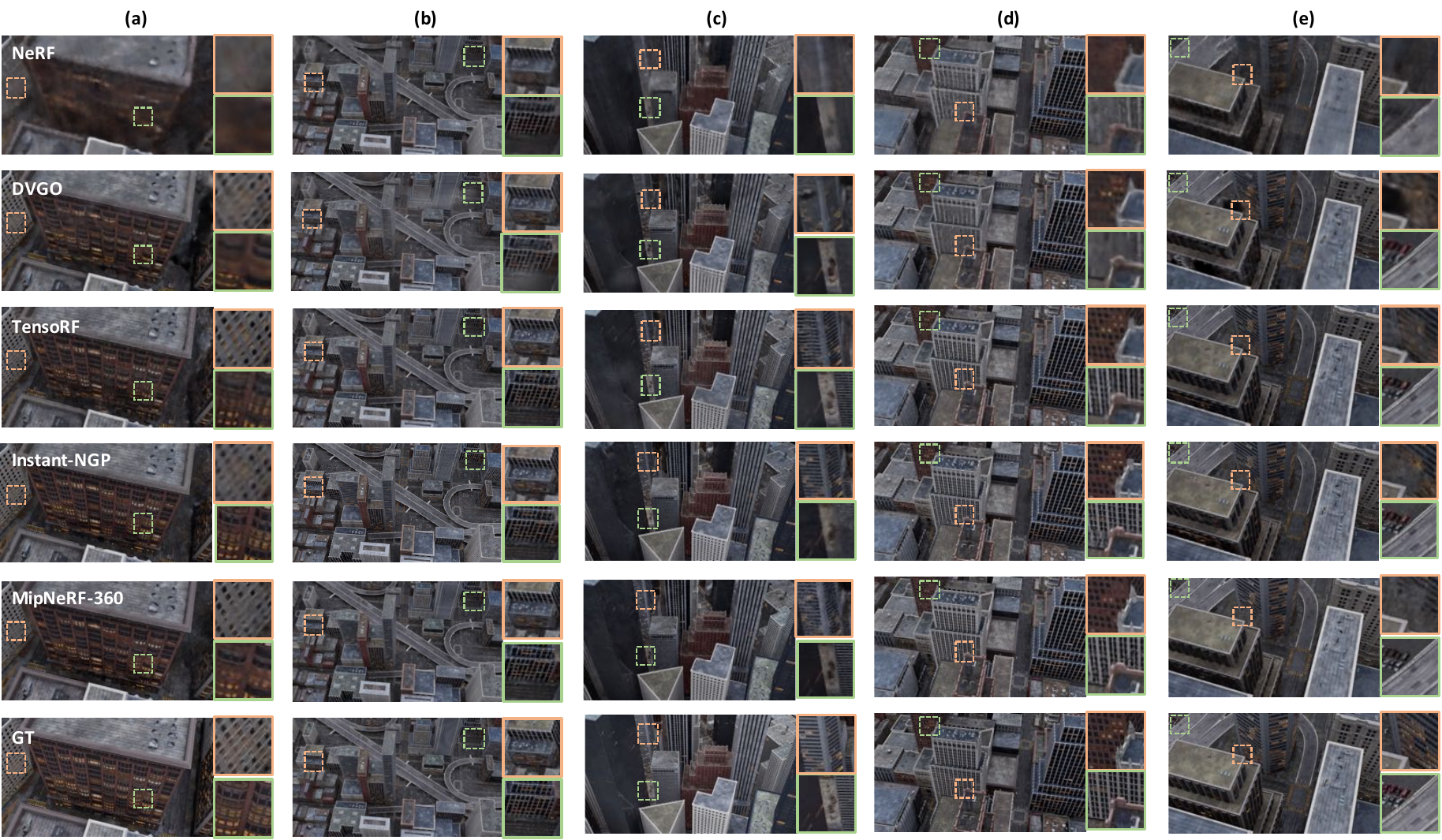}
\end{center}
   \caption{Visualization of novel view synthesis results of previous representative large-scale neural rendering methods on the aerial data of our {\em MatrixCity} dataset.}
\label{fig:high_result}
\end{figure*}
\section{Experiments}

\begin{figure*}[t]
\begin{center}
\includegraphics[width=1.0\textwidth]{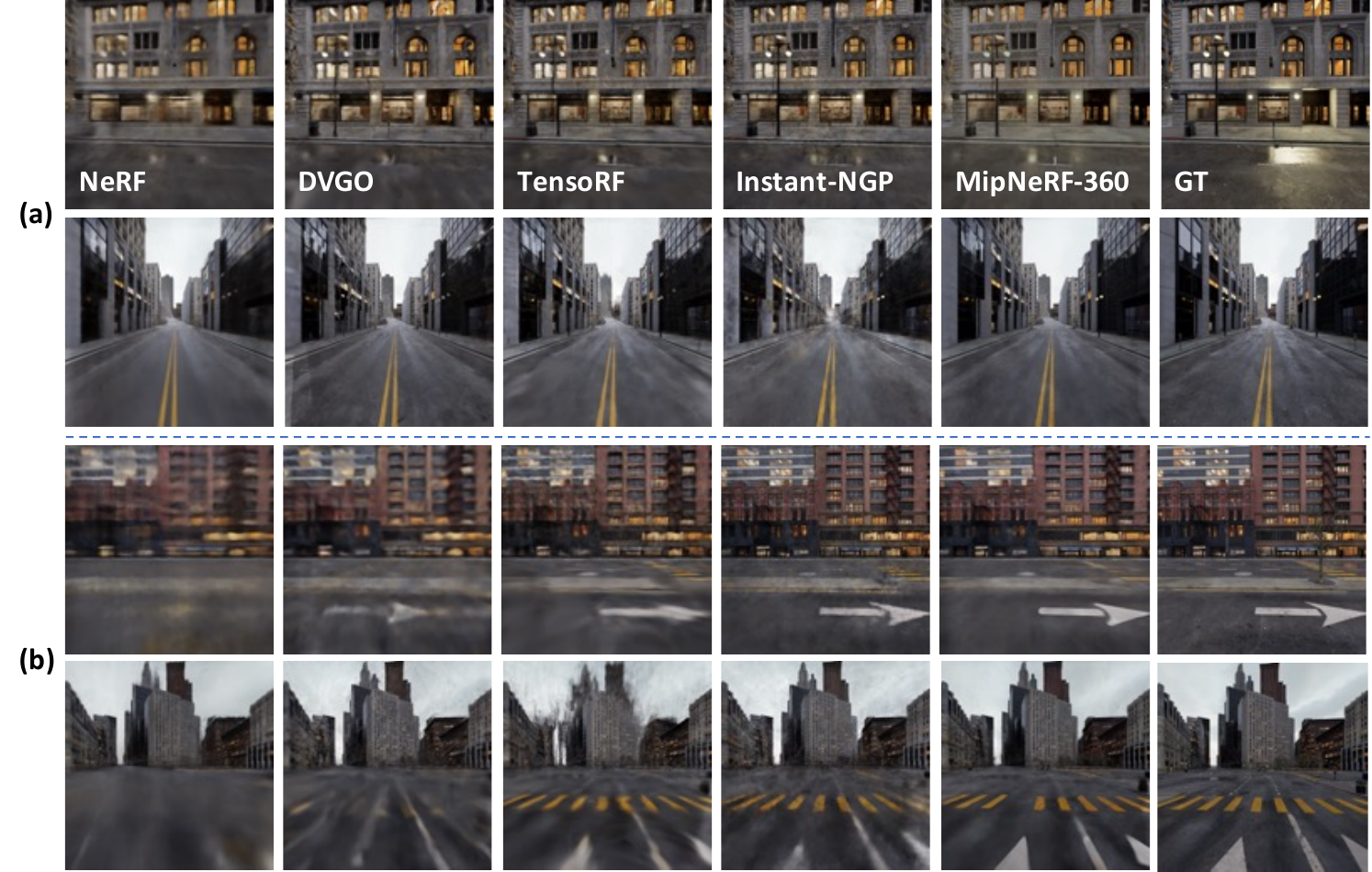}
\end{center}
   \caption{Visualization of novel view synthesis of city-scale neural rendering methods on (a) Block\_Small and (b) Block\_A of street-view data. MLP-based NeRF methods suffer from capacity issues while grid-based baselines shows severe artifacts.}
\label{fig:visual_road}
\end{figure*}

In this section, we mainly investigate the quality of novel view rendering, and reveal the challenges of adapting existing SOTA methods on this task. Additional studies (\eg, dynamics scenes, lighting control, \etc) are provided in the Appendix D.

\subsection{Datasets and Metrics} 
\noindent\textbf{MatrixCity benchmark.} The MatrixCity dataset contains two city maps: Small City and Big City. According to the common practice in surveying and mapping that adjacent images should have an overlap of 70\%-80\%, we set a camera capture location every 40 m for aerial data collection and 5 m for street data collection. Small City includes 6k aerial images and 30k street-level images, while Big City has 60 k aerial images and 286 k street-level images. Note that we remove the aerial images that look outside the map boundary manually. Also, we remove the street images that look straight down following nerfstudio~\cite{nerfstudio}, which crops the bottom 20\% of the 360 images to reduce useless information. The ratio of training set to testing set is 8:1. To ensure both completeness of training perspectives and generalization ability in testing, test set is collected separately with no location overlap with the training set. For aerial data, the yaw direction randoms from $0\degree$ to $360\degree$ and the pitch direction randoms from $-60\degree$ to $-45\degree$, and every camera location captures 1 image. For street data, the yaw direction randoms from $0\degree$ to $90\degree$ and every camera location captures 5 images, whose pitch and roll direction keep the same with the training set. Since the street data contains more details, we also ablate the street data collection density in Table~\ref{tab:road_density}, which demonstrates that grid-based method is more sensitive to data density than MLP-based NeRF method. Additionally, we provide a super dense version street data with 135k for Small City with 1 m interval. For demonstrative purpose, we conduct experiments on the Small City in this stage, where the interval between adjacent frames is 5 m for street data. We will release the data splits of the following sections.

\smallskip
\noindent\textbf{Evaluation metric.} We evaluate the rendering performance of each baseline method based on PSNR(Peak Signal-to-Noise Ratio), SSIM(Structural Similarity)~\cite{DBLP:journals/tip/WangBSS04} and the VGG implementation of LPIPS~\cite{DBLP:conf/cvpr/ZhangIESW18}. And we also use mean angular error (MAE) and mean squared error (MSE) to evaluate estimated normal vectors and depth map, respectively.

\begin{table*}[htbp]
\centering
\resizebox{\linewidth}{!}{
\begin{tabular}{c|ccc|ccc|ccc|ccc|ccc}
\toprule
                                 & \multicolumn{3}{c|}{ \textbf{NeRF ~\cite{DBLP:conf/eccv/MildenhallSTBRN20}}}                                            & \multicolumn{3}{c|}{\textbf{DVGO} ~\cite{DBLP:conf/cvpr/0004SC22}}                                            & \multicolumn{3}{c|}{\textbf{TensoRF} ~\cite{DBLP:conf/eccv/ChenXGYS22}}                                                                & \multicolumn{3}{c|}{\textbf{Instant-NGP} ~\cite{DBLP:journals/tog/MullerESK22}} & \multicolumn{3}{c}{ \textbf{MipNeRF-360} ~\cite{DBLP:conf/cvpr/BarronMVSH22}} \\
\multirow{-2}{*}{Block}        & PSNR $\uparrow$                        & SSIM  $\uparrow$                       & LPIPS $\downarrow$                       & PSNR $\uparrow$                        & SSIM  $\uparrow$                       & LPIPS $\downarrow$                       & PSNR $\uparrow$                       & SSIM  $\uparrow$                       & LPIPS  $\downarrow$                      & PSNR  $\uparrow$           & SSIM  $\uparrow$           & LPIPS  $\downarrow$          & PSNR $\uparrow$            & SSIM  $\uparrow$           & LPIPS $\downarrow$           \\ \midrule
Block\_A  & 20.12 & 0.601 & 0.626 & 20.47 & 0.617 & 0.604 & 20.93 & 0.643 & 0.577 & \underline{21.96}           & \underline{0.712}            & \underline{0.493}            & \textbf{22.00}            & \textbf{0.717}            & \textbf{0.488}            \\
Block\_Small & 22.15 & 0.678 & 0.511 & 22.10 & 0.711 & 0.454 & \underline{22.95} & 0.741 & 0.445 & 22.84           & \underline{0.745}            & \underline{0.408}            & \textbf{24.47}            & \textbf{0.827}            & \textbf{0.297}            \\
Overall                             & 21.14                        & 0.640                        & 0.569                        & 21.29                        & 0.664                        & 0.529                        & 21.94                        & 0.692                        & 0.511                        & \underline{22.40}            & \underline{0.729}            & \underline{0.451}            & \textbf{23.24}            & \textbf{0.772}            & \textbf{0.393}   \\        
\bottomrule
\end{tabular}
}

\caption{Performance comparison of representative neural rendering methods on the street data of our {\em MatrixCity} benchmark.}
 \label{tab:street_results}%
\end{table*}

\subsection{Baselines}

We aim to test the performance of current neural rendering methods on the MatrixCity dataset to explore the challenges for city-scale neural rendering. To achieve this, we choose five widely recognized methods: NeRF~\cite{DBLP:conf/eccv/MildenhallSTBRN20}, DVGO~\cite{DBLP:conf/cvpr/0004SC22}, Instant-NGP~\cite{DBLP:journals/tog/MullerESK22}, TensoRF~\cite{DBLP:conf/eccv/ChenXGYS22} and MipNeRF-360~\cite{DBLP:conf/cvpr/BarronMVSH22}. Note that we all use the official implementation of these baselines except NeRF and Instant-NGP. For NeRF we use the widely recognized Pytorch version ~\cite{lin2020nerfpytorch}. And for Instant-NGP, we use the open-source version ~\cite{ngp-pl}. We find that ngp-pl~\cite{ngp-pl} generally performs better than torch-ngp~\cite{torch-ngp}. To address the challenge of increasingly intricate urban content, we recognize the limited capacity of the original baseline models. So we increase the number of parameters to handle more complex urban environments. Specific details regarding these parameter increases can be found in the Appendix A.

\subsection{Neural Rendering on Aerial Data}

Due to the limitations of current methods and models, it is impractical to use a single model to represent an entire map. Therefore, we divide the map into five blocks based on building height and coverage area. Each block covers a roughly homogeneous area, where buildings within each block have similar heights. Our results, shown in Table~\ref{tab:high_results}, indicate that MipNeRF-360~\cite{DBLP:conf/cvpr/BarronMVSH22} and Instant-NGP~\cite{DBLP:journals/tog/MullerESK22} perform better, while NeRF~\cite{DBLP:conf/eccv/MildenhallSTBRN20} performs the worst. This indicates that grid-based methods and MLP-based NeRF methods both can model the block-size aerial data well. Despite scaling up the NeRF model significantly, its ability to model large-scale scenes remains limited, as illustrated in Figure~\ref{fig:high_result}. Additionally, we found that the high-rise area is more challenging to model than the floor area. In the high-rise area, there are numerous occlusions between the buildings, which is a significant challenge for aerial data modeling. From Figure~\ref{fig:high_result}, we can observe that current methods still struggle to accurately model small objects and reflective buildings.

\subsection{Neural Rendering on Street Data}

\begin{figure*}[h]
\begin{center}
\includegraphics[width=1.0\textwidth]{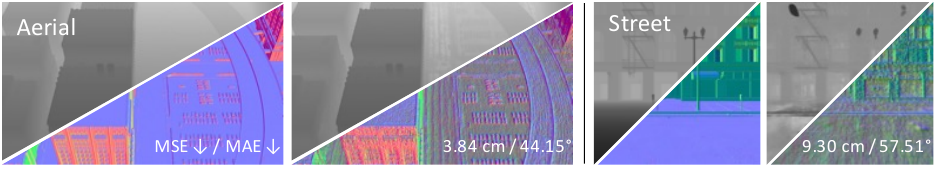}
\end{center}
\caption{Visualization of the depth and normal results of MipNeRF-360 on aerial and street views.}
\label{fig:depth_normal}
\end{figure*}

\begin{table}[]
\centering
\resizebox{\linewidth}{!}{
\begin{tabular}{c|ccc|ccc}
\toprule
  & \multicolumn{3}{c|}{\textbf{Instant-NGP} ~\cite{DBLP:journals/tog/MullerESK22}}                                                                & \multicolumn{3}{c}{{\textbf{MipNeRF-360} ~\cite{DBLP:conf/cvpr/BarronMVSH22}}}                                     \\
\multirow{-2}{*}{Density}   & PSNR  $\uparrow$                       & SSIM  $\uparrow$                       & LPIPS  $\downarrow$                      & PSNR $\uparrow$                        & SSIM  $\uparrow$                       & LPIPS $\downarrow$                       \\ \midrule
5.0 m & {21.436} & {0.733} & {0.402} & {27.75} & {0.866} & {0.2956} \\
3.6 m & {24.978} & {0.803} & {0.350} & {29.366} & {0.884} & {0.280} \\
2.0 m  & {30.025} & {0.885} & {0.235} & {31.420} & {0.901} & {0.265} \\
1.0 m & {32.444} & {0.912} & {0.211} & {31.858} & {0.905} & {0.263} \\
0.5 m & {32.999} & {0.921} & {0.202} & {32.210} & {0.907} & {0.261} \\
\bottomrule
\end{tabular}
 }
\caption{Ablation on the density of street data collection on our {\em MatrixCity} benchmark.}
\vspace{-8pt}

\label{tab:road_density}
\end{table}

\begin{figure}[ht]
\begin{center}
\includegraphics[width=0.5\textwidth]{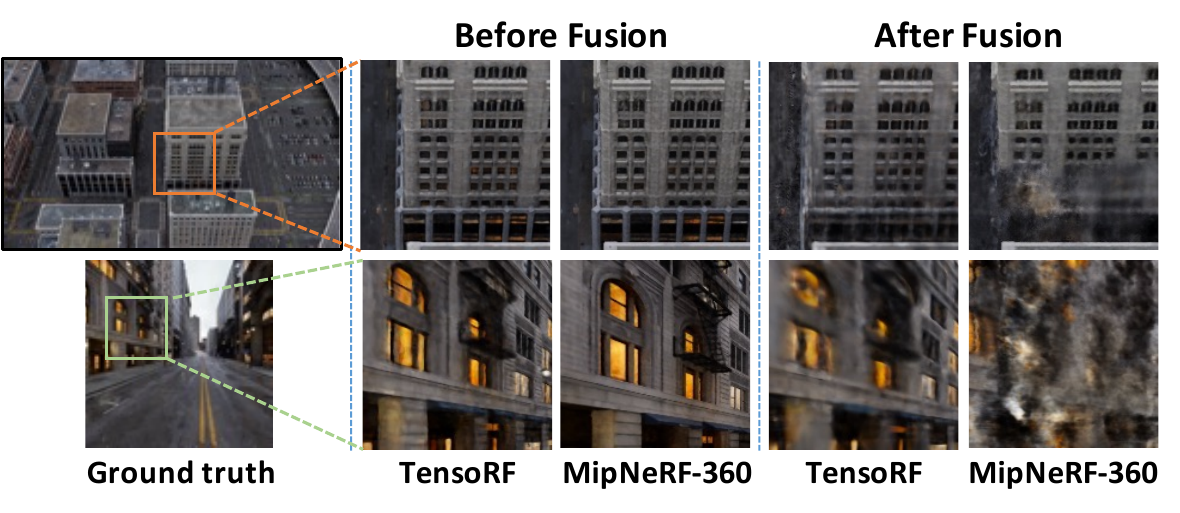}
\end{center}
\vspace{-5pt}
   \caption{Visualization of neural rendered results on aerial and street views before and after fusion of two types of views. Streets views are generally harder than aerial views to deliver high-quality rendering results, with notable floating artifacts, where the model get easily overfitting to the training views with cheated geometry. The naive joint training on the fused data downgrades the quality. }
\label{fig:visual_fusion}
\end{figure}

We first run all these baselines on the street data of Block\_A and find that all the methods perform much worse than the results of training with aerial data, especially for the grid-based methods, as shown in Table~\ref{tab:street_results}. 
Street data contains much more details than aerial data, and it is harder to achieve high-quality results on street data, which is also demonstrated in Figure~\ref{fig:depth_normal}. 
Thus we conclude that modeling the street data of a block-size area in a single model is not reasonable and filter a crossroad data to test current methods, called Block\_Small. Analyzed the results on Block\_Small, we find that the MLP-based NeRF methods perform better than the grid-based methods, which is also demonstrated in Figure~\ref{fig:visual_road}. The Block\_Small can also be seen as a 360 unbounded scenes with a distant background. Figure~\ref{fig:visual_road} shows that MipNeRF-360 can alleviate this problem to some extent. However, the reflective parts and fine-grained architectures are still not well reconstructed.

\subsection{Neural Rendering on Joint Types of Data}
\begin{table}[]
\centering
\resizebox{\linewidth}{!}{
\begin{tabular}{c|ccc|ccc}
\toprule
  & \multicolumn{3}{c|}{\textbf{TensoRF} ~\cite{DBLP:conf/eccv/ChenXGYS22}}                                                                & \multicolumn{3}{c}{{\textbf{MipNeRF-360} ~\cite{DBLP:conf/cvpr/BarronMVSH22}}}                                     \\
\multirow{-2}{*}{Data Type}   & PSNR  $\uparrow$                       & SSIM  $\uparrow$                       & LPIPS  $\downarrow$                      & PSNR $\uparrow$                        & SSIM  $\uparrow$                       & LPIPS $\downarrow$                       \\ \midrule
{Aerial} & {27.26} & {0.829} & {0.231} & {28.37} & {0.855} & {0.197} \\
{Streat} & {22.10} & {0.727} & {0.449} & {23.05} & {0.805} & {0.312} \\
Fusion                  & {21.44} & {0.656} & {0.504} & {17.07} & {0.470} & {0.600} \\
\bottomrule
\end{tabular}}
\caption{Ablation on the fusion of aerial and street data on our {\em MatrixCity} benchmark.}
\vspace{-1em}
\label{tab:road+high_benchmark}
\end{table}
The major motivation to fuse data from both aerial and street view is to provide content information at different granularity. While aerial views are generally easier to train with less geometry ambiguities, it lacks many details on the near-ground, which are critical to deliver immersive experience for exploring a city. On the other hand, street-view images often only offer partial information about the scene revealing the local contents, which are sensitive to overfit to training views. 
We therefore explore to train the aerial and street data together, which cover the same area, aiming the leverage the advantage of two sources of data to ensure wide coverage as well as fine details. However, according to Table~\ref{tab:road+high_benchmark}, we find that the performance of TensoRF~\cite{DBLP:conf/eccv/ChenXGYS22} and MipNeRF-360~\cite{DBLP:conf/cvpr/BarronMVSH22} both got worse after naively fusing the aerial and street data to train together. As shown in the Figure ~\ref{fig:visual_fusion}, the ground part of the aerial view becomes dirty after training with the street data for both methods. For the street view, the foreground of MipNeRF-360 becomes worse. We analyze that due to the significant difference in the level of details between street-level and aerial imagery, as well as the large disparity in distance from the foreground, it is challenging to train models simply by utilizing both types of data. We need to further investigate how algorithms can effectively utilize both the geometric information from aerial imagery and the detailed information from street-level imagery, such as finetuning, progressive training, separate group of hyperparamters, \etc.

\section{Conclusion}
In this paper, we propose {\em MatrixCity}, a high-quality and city-scale benchmark with diverse, controllable and realistic data collected from the powerful Unreal Engine 5. Additional information like depth and normal are also collected with minimum extra cost in our {\em MatrixCity} dataset, enables other potential tasks and applications like depth estimation and inverse rendering. On top of {\em MatrixCity}, we have empirically investigated representative methods on two types of data independently and the fusion of both aerial-view and street-view data. We hope these efforts could facilitate new advances in the field of city-scale neural rendering. 

\noindent\textbf{Acknowledgment} This project is funded in part by Shanghai AI Laboratory (P23KS00020, 2022ZD0160201), CUHK Interdisciplinary AI Research Institute, and the Centre for Perceptual and Interactive Intelligence (CPIl) Ltd under the Innovation and Technology Commission (ITC)'s InnoHK. We would like to thank Haiyi Mei and Lei Yang for their invaluable help and discussions for the plug-in development, Jiamian Yan, Bin Wang and Conghui He for their contributions on the street data annotations of Big City.

\clearpage
{\small
\bibliographystyle{ieee_fullname}
\bibliography{egbib}
}
\begin{table*}[htbp]
\small
\centering
\resizebox{0.8\linewidth}{!}{
\begin{tabular}{c|cccccccccc}
\toprule
Block & 1 & 2 & 3 & 4 & 5 & 6 & 7 & 8 & 9 & 10 \\ \midrule
Height   & 150 m & 150 m & 300 m  & 500 m & 450 m & 450 m  & 350 m & 350 m & 250 m & 200m \\
\bottomrule
\end{tabular}
}
\caption{Heights of different aerial block splits for the \emph{Small City} data collection.}
\label{tab:high_block}
\end{table*}

\section*{Appendix A: Method Details}

\noindent\textbf{NeRF}~\cite{DBLP:conf/eccv/MildenhallSTBRN20} is the pioneering method that represents a scene implicitly using a fully-connected MLP network. It can synthesize high-quality novel views through volume rendering. We utilize 12 layers with the width 512 for both the coarse and fine networks. The 3D location of the maximum frequency for positional encoding is set to 16 instead of 10. Additionally, we incorporate three skip connections that concatenate the input to the activations of the 4th, 8th, and 10th layers.

\noindent\textbf{DVGO}~\cite{DBLP:conf/cvpr/0004SC22} represents the scene by means of optimizing dense grids, comprising of density grids for scene geometry and feature grids with a shallow MLP network for view-dependent appearance in detail. We removes the coarse geometry search step because the view-count-based learning rate prior is ineffective for urban scenes. This is because the edge views are often underrepresented, making it challenging to optimize the algorithm. Furthermore, we set the number of voxels in the fine stage to $500^3$ to ensure a fair comparison with other methods.

\noindent\textbf{Instant-NGP}~\cite{DBLP:journals/tog/MullerESK22} introduces a multi-resolution hash encoding structure, which guarantees both efficiency and accuracy in the modeling process. We increase the number of parameters for the hash encoding. Specifically, we utilize 16 levels for the hash encoding, with each individual hash table containing $2^{22}$ entries. The resolution ranged from 16 to 65536.

\noindent\textbf{TensoRF}~\cite{DBLP:conf/eccv/ChenXGYS22} models the scenes as a 4D tensor, consisting of 3D voxel grids and a multi-channel feature. These tensors are then decomposed into compact vector and matrix factors for more efficient processing. In our paper, we adapt the grid resolution to $500^3$ to better model the large-scale scene and upsample the grids resolution at the 2000th, 3000th, 4000th, 5500th, 7000th, 10000th, 12000th, 14000th iterations. Additionally, we discover that the softplus activation function is unstable for large-scale urban scenes and thus replace it with ReLU.

\noindent\textbf{MipNeRF-360}~\cite{DBLP:conf/cvpr/BarronMVSH22} employs a non-linear parameterization technique, coarse-to-fine online distillation, and a distortion loss function to address unbounded artifacts. We normalize the camera poses of the trained images into a unit sphere for the contraction operation. We utilize 4 layers-MLP with the width 256 for the propose network and 8 layers-MLP with the width 1024 for the nerf network.

For all the experiments in our paper, we resize the 1080P images into $540\times960$ for training and testing. Each single model except MipNeRF-360 is trained on a single Nvidia A100 GPU device for around 0.5-30 hours. We use 4 Nvidia A100 GPU to train MipNeRF-360 for around 10 hours.

\section*{Appendix B: More Results on Aerial Data}

\begin{table*}[]
\centering
\resizebox{\linewidth}{!}{
\begin{tabular}{c|ccc|ccc|ccc|ccc}
\toprule
  & \multicolumn{3}{c|}{\textbf{NeRF}~\cite{DBLP:conf/eccv/MildenhallSTBRN20}}                                                                  & \multicolumn{3}{c|}{{\textbf{TensoRF} ~\cite{DBLP:conf/eccv/ChenXGYS22}}}  &
  \multicolumn{3}{c|}{{\textbf{Instant-NGP} ~\cite{DBLP:journals/tog/MullerESK22}}} & \multicolumn{3}{c}{{\textbf{MipNeRF-360} ~\cite{DBLP:conf/cvpr/BarronMVSH22}}}\\
\multirow{-2}{*}{Data Type}   & PSNR  $\uparrow$                       & SSIM  $\uparrow$                       & LPIPS  $\downarrow$                      & PSNR $\uparrow$                        & SSIM  $\uparrow$                       & LPIPS $\downarrow$  & PSNR $\uparrow$                        & SSIM  $\uparrow$                       & LPIPS $\downarrow$  & PSNR $\uparrow$                        & SSIM  $\uparrow$                       & LPIPS $\downarrow$                    \\ \midrule
{Multi-model} & {22.97} & {0.589} & {0.548}  & {25.13} & {0.762} & {0.396} & \underline{25.69} & \textbf{0.800} & \textbf{0.326} & \textbf{25.73} & \underline{0.779} & \underline{0.360}\\
{Single-model} & {21.69} & {0.510} & {0.638}  & \underline{25.09} & \textbf{0.752} & \textbf{0.370} & \textbf{25.46} & \underline{0.751} & \underline{0.402} & {24.41} & {0.689} & {0.469}\\
\bottomrule
\end{tabular}}
\caption{Performance comparison of different kinds of methods on the aerial data of our $MatrixCity$ benchmark with the multi-model and the single model settings. Note that the multi-model represents that we divide the small city into five blocks and train a separate model for each block. The single-model represents that we train a single model for the whole small city.}
\label{tab:whole_map}
\end{table*}

\begin{figure*}[t]
\begin{center}
\includegraphics[width=1.0\textwidth]{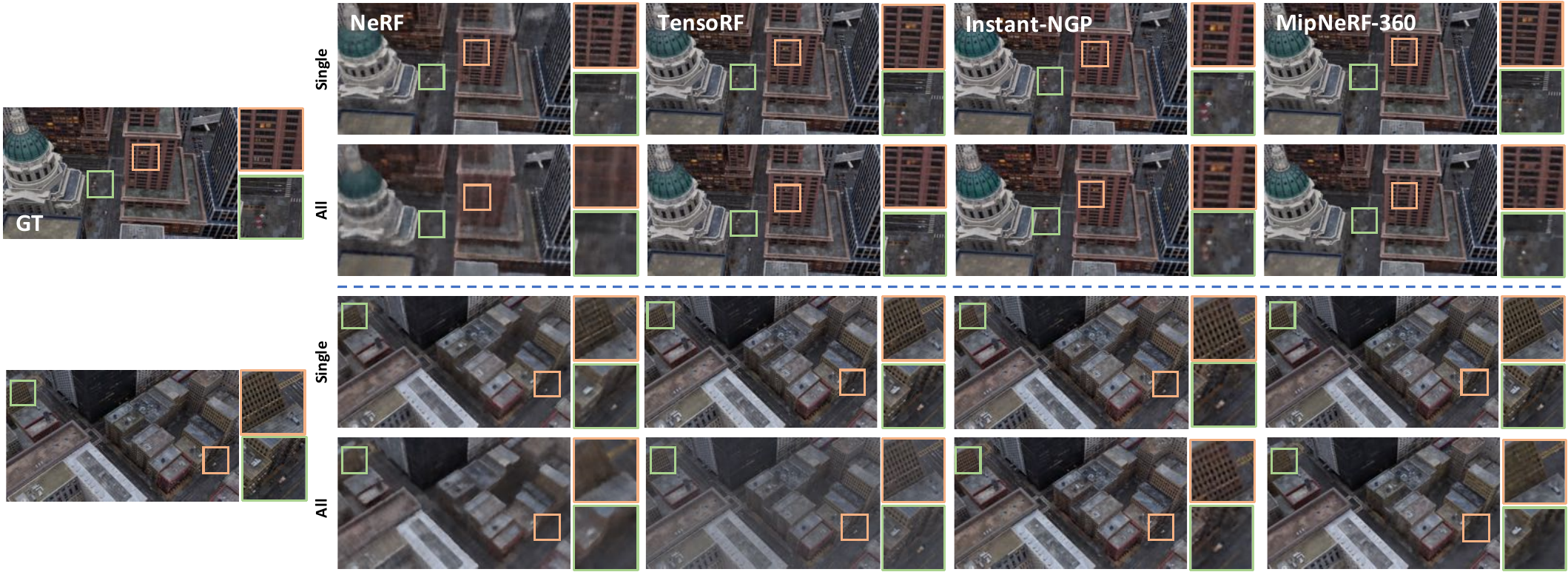}
\end{center}
   \caption{Visualization of novel view synthesis results of different kinds of methods on the aerial data of our $MatrixCity$ benchmark with the multi-model and the single model settings. Note that the multi-model represents that we divide the small city into five blocks and train a separate model for each block. The single-model represents that we train a single model for the whole small city.}
\label{fig:single+multi}
\end{figure*}

To investigate the characteristics of the grid-based methods and MLP-based nerf methods for large area modeling, we conduct experiments using aerial data of an entire small city. We increase the capacity of the MipNeRF-360 by extending its nerf network to 12 layers, and enlarge the grid resolution of TensoRF to $1500^3$. As shown in Table~\ref{tab:whole_map}, the MLP-based nerf methods, NeRF and MipNeRF-360, exhibit a significant decline in performance when modeling larger areas, although the model has been enlarged. Unlike MLP-based methods that utilize continuous networks, the grid-based approach of TensoRF and Instant-NGP utilizes discrete grids. By increasing the grid resolution, the model exhibits almost no drop in performance when modeling larger areas. As illustrated in Figure~\ref{fig:single+multi}, the learning performance for details significantly diminished for both NeRF and MipNeRF-360 after expanding the area size, despite the accompanying increase in model size. For TensoRF and Instant-NGP, expanding the area size with increasing the grid resolution in proportion does not significantly affect the learning of details. However, as shown int the second row of Figure~\ref{fig:single+multi}, training high-rise buildings and low-rise buildings together with a shared bbox can lead to the air above the low-rise buildings being relatively dirty, which is a point that needs to be optimized in future work.

\section*{Appendix C: More results on Street Data}

We notice that BlockNeRF~\cite{DBLP:conf/cvpr/TancikCYPMSBK22} uses a small block size~($\sim 0.031km^2$), so we experiment on a similar setting in Table~\ref{tab:ablation}. We also ablate block size and model capacity on street views in the table below. Note that the grid resolution of TensoRF is $300/500/800$, and the network width of MipNeRF-360 is $256/512/1024$ for small/medium/large model size. We use the same number of rendering samples as the original paper for all methods. 

\begin{table}[h]
\centering
\resizebox{\linewidth}{!}{
\begin{tabular}{c|c|c|ccc|ccc}
\toprule
    & & & \multicolumn{3}{c|}{{\textbf{TensoRF}}}  & \multicolumn{3}{c}{{\textbf{MipNeRF-360}}}                                    \\
\multirow{-2}{*}{Blocks} & \multirow{-2}{*}{Size}& \multirow{-2}{*}{Model} & PSNR $\uparrow$                        & SSIM  $\uparrow$                       & LPIPS $\downarrow$ & PSNR $\uparrow$                        & SSIM  $\uparrow$                       & LPIPS $\downarrow$ \\ \midrule
{1} & $0.540km^2$ & Large&21.23 & 0.663 & 0.556 & 22.00 & 0.717 & 0.488 \\
{4} & $0.135km^2$& Large& 22.14 & 0.710& 0.472& 25.83 & 0.802 & 0.381 \\
{8} & $0.068km^2$ & Large & 23.47 & 0.751 & 0.432& 26.44 & 0.835 & 0.320 \\
{16} & $0.034km^2$ & Large& \underline{24.24} & \underline{0.798} & \underline{0.376} & \textbf{26.67} & \textbf{0.858} & \textbf{0.274} \\
{16} & $0.034km^2$ & Medium & 23.60 & 0.767& 0.417& 26.03 & 0.822 & 0.345 \\
{16} & $0.034km^2$ & Small & 22.96 & 0.727 & 0.469 & 24.64& 0.763 & 0.432 \\
\bottomrule
\end{tabular}
}
\caption{Ablation on block size and model capacity on the street data of Small City.}
\label{tab:ablation}
\end{table}

\section*{Appendix D: Extension Study}
To delve deeper into the challenges of novel view synthesis in the real-world urban scenes, we design two sets of experiments that involve changing lighting conditions and dynamic scenes . 

\subsection*{D.1. Illumination}

\begin{figure*}[t]
\begin{center}
\includegraphics[width=1.0\textwidth]{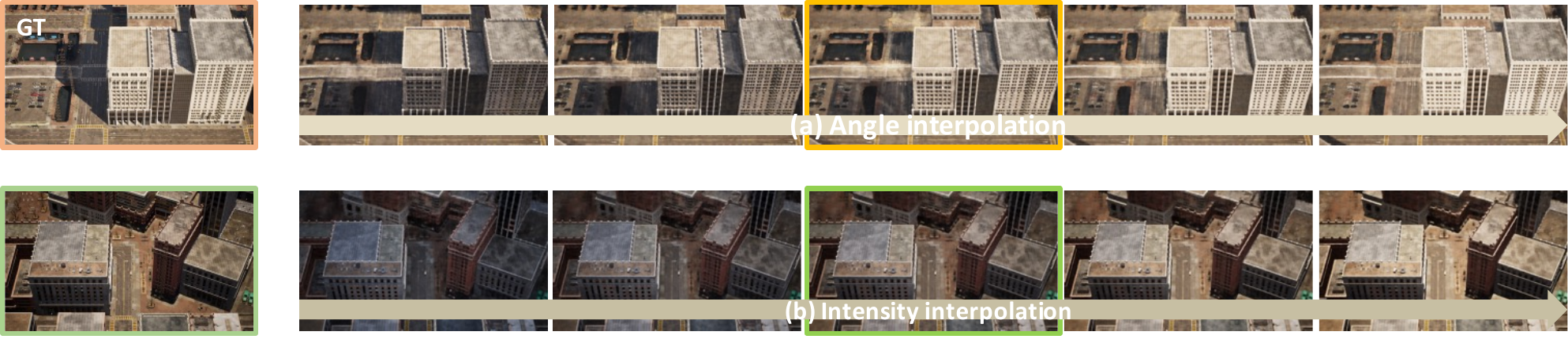}
\end{center}
   \caption{Visualization of the interpolation result of light angle (1st row) and light intensity (2nd row). In the first row, the angle of the first image is 0 degree and the angle of the last image is 90 degree. The middle image in the orange box is the interpolated 45 degree's result. In the second row, the intensity of the first image is 1000 and the intensity of the last image is 3000. The middle image in the green box is the interpolated 2000 intensity's result.}
\label{fig:intensity&direction}
\end{figure*}

In the real-world scenes, the intensity and direction of the light change throughout the day. We decouple these two dimensions to explore the challenges involved. For the light direction, we collect data with different light angle from 0 degree to 90 degree with a interval of 5 degree. The first row of Figure~\ref{fig:intensity&direction} shows the interpolation result of the direction of light from the first image (0 degree) to the last image (90 degree). We can see that the shadow only appears lighter, but it does not capture the essential relationship between the interaction of light and the building structure. For light intensity, we collect data with three different light intensity. In the second row of Figure~\ref{fig:intensity&direction}, we can observe a reasonable and continuous change from the first image to the last image. However, there still exists a visual discrepancy between the interpolated images and the ground-truth images. This implies that the decoupling of light intensity and scene's color is not well executed.

\subsection*{D.2. Dynamic}
\begin{figure}[htbp]
\begin{center}
\includegraphics[width=0.3\textwidth]{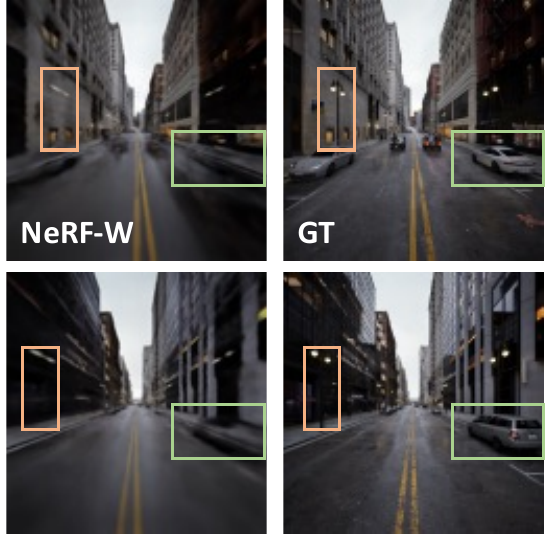}
\end{center}
\vspace{-1em}
   \caption{Visualization of the novel view synthesis results of the stationary scene with moving cars and people.}
\label{fig:dynamic}
\end{figure}

We gather a collection of street view images featuring moving people and traffic, aiming to model the stationary buildings and filter out the dynamic traffic using NeRF-W~\cite{DBLP:conf/cvpr/Martin-BruallaR21}, a method that can distinguish between dynamic objects and static buildings. However, as shown in Figure~\ref{fig:dynamic}, we encounter a setback during the decomposing process; some stationary objects such as parked cars and streetlights are accidentally filtered out, which is unacceptable.

\newpage

\end{document}